\documentclass[letterpaper, 10pt, conference]{ieeeconf}  % Comment this line out if you need a4paper

\IEEEoverridecommandlockouts                              % This command is only needed if
\usepackage{amsthm}
\usepackage{times}
\usepackage{multicol}
\usepackage[bookmarks=true]{hyperref}
\usepackage{xcolor}
\usepackage{hyperref}
\usepackage{amsmath, amssymb}
\usepackage{amsfonts}
\usepackage{graphicx}
\usepackage{siunitx}
\usepackage{standalone}
\usepackage{booktabs}
\usepackage[ruled,vlined,linesnumbered,noend]{algorithm2e}
\usepackage{mdframed}
\usepackage{fancyvrb,multirow}
\usepackage{soul}
\usepackage{dsfont,mathabx}
\usepackage{array, booktabs}
\usepackage{subfigure}
\usepackage{booktabs}
\usepackage{makecell}
\usepackage{threeparttable}

\newtheorem{theorem}{Theorem}
\newtheorem{proposition}{Proposition}

\theoremstyle{definition}

\newtheorem{definition}{Definition}

\newtheorem{problem}{Problem}

\title{\LARGE \bf
  Combinatorial-hybrid Optimization for Multi-agent Systems\\
  under Collaborative Tasks
}

\author{Zili Tang, Junfeng Chen and Meng Guo% <-this % stops a space
\thanks{The authors are with the College of Engineering, Peking University, Beijing 100871, China.
    This work was supported by the National Natural Science Foundation
    of China (NSFC) under grants 62203017, T2121002, U2241214;
    and by the Fundamental Research Funds for the central universities.
    Contact: {\tt\small tang.zili, chenjunfeng, meng.guo@pku.edu.cn}.}% <-this % stops a space
}

\begin{document}
\maketitle
\thispagestyle{empty}
\pagestyle{empty}

%%========================================

%%========================================
\begin{abstract}
  Multi-agent systems can be extremely efficient when working concurrently and collaboratively,
  e.g., for transportation, maintenance, search and rescue.
  Coordination of such teams often involves two aspects:
  (i) selecting appropriate sub-teams for different tasks;
  (ii) designing collaborative control strategies to execute these tasks.
  The former aspect can be combinatorial w.r.t. the team size,
  while the latter requires optimization over joint state-spaces
  under geometric and dynamic constraints.
  Existing work often tackles one aspect by assuming the other is given,
  while ignoring their close dependency.
  This work formulates such problems as combinatorial-hybrid optimizations (CHO),
  where both the discrete modes of collaboration and the continuous control parameters are
  optimized simultaneously and iteratively.
  The proposed framework consists of two interleaved layers:
  the dynamic formation of task coalitions and the hybrid optimization
  of collaborative behaviors.
  Overall feasibility and costs of different coalitions performing various tasks
  are approximated at different granularities to improve the computational efficiency.
  At last, a Nash-stable strategy for both task assignment and execution
  is derived with provable guarantee on the feasibility and quality.
  Two non-trivial applications of collaborative transportation and dynamic capture
  are studied against several baselines.
\end{abstract}

%%========================================
%==============================
\section{Introduction}\label{sec:intro}
Fleets of heterogeneous and autonomous robots are depolyed nowadays to accomplish tasks
that are otherwise too inefficient or even infeasible for a single robot,
e.g., collaborative transportation~\cite{bock2007construction, krizmancic2020cooperative},
dynamic capture~\cite{pierson2016intercepting}
and surveillance~\cite{khan2016cooperative}.
Both the overall efficiency and capability of the team are significantly improved
by allowing the team of robots to act concurrently and collaboratively.
Two aspects are often involved for the coordination of such teams.
On the one hand, the given set of tasks could be accomplished by various
subgroups of the team,
however at drastically different costs~\cite{torreno2017cooperative,gini2017multi,massin2017coalition}.
For instance, three agents can detect and capture a dynamic target much faster than one agent
(if possible at all), while five agents might be redundant in certain scenarios.
Thus, an appropriate task assignment is crucial for the overall performance,
which unfortunately often has a complexity combinatorial to the number of agents and tasks.
On the other hand, given certain assignments, how each subgroup executes the assigned task often
boils down to an optimal control problem~\cite{krizmancic2020cooperative, weintraub2020introduction}, i.e., how to actuate the agents collaboratively
to minimize the cost associated with a task, under the dynamic
and geometric constraints~\cite{gini2017multi, tordesillas2021mader}.
Exact optimization has a high complexity due to the long horizon
and the high dimension of joint-state-control space of all collaborators.
Lastly, there is often a ``chicken or the egg'' dilemma w.r.t. these two aspects~\cite{makkapati2019optimal,hartmann2020robust}.
Namely, the task assignment relies on the optimal control to
evaluate feasibility and actual cost,
while the optimal control problem requires certain assignments as inputs.
Solving first the control problems for \emph{all} possible assignments
and then the task assignment problem is mostly intractable,
as it multiplies the complexity of both aspects.

%==============================
\subsection{Related Work}\label{subsec:intro-related}

% task planning and motion planning separately
Task planning for multi-agent systems
refers to the process of first decomposing this task into sub-tasks
and then assigning them to the team,
see~\cite{torreno2017cooperative,gini2017multi,khamis2015multi} for comprehensive surveys.
Different optimization criteria can be chosen, typically:
MinSUM that minimizes the sum of agent costs over all
agents~\cite{gini2017multi, fukasawa2006robust};
and MinMAX that minimizes the maximum cost of all agents~\cite{nunes2015multi}.
Well-known problems include
the classic one-to-one assignment problem~\cite{jonker1986improving},
the multi-Vehicle routing problem~\cite{gini2017multi, khamis2015multi},
and the coalition formation problem~\cite{massin2017coalition,apt2009generic}.
Representative methods include the Hungarian method~\cite{jonker1986improving},
the mixed integer linear programming (MILP)~\cite{torreno2017cooperative},
the search-based methods~\cite{fukasawa2006robust};
and the market-based methods~\cite{khamis2015multi}.
However, this body of literature normally assumes a static and known table of task-per-agent costs,
which is not always available or even \emph{invalid} for collaborative tasks,
as the benefit of one agent joins a task depends on which other agents also participate.

Motion planning for multi-agent systems refers to the design of control strategies
for each agent to accomplish a given task.
One common task is the collaborative navigation where each agent navigates
to its goal position while avoiding collision with other agents or obstacles,
see~\cite{soria2021distributed, tordesillas2021mader}.
Other tasks such as formation, flocking are also studied under various constraints,
see~\cite{sun2016optimal}.
Another relevant task is the collaborative load
transportation~\cite{tuci2018cooperative, alevizos2022bounded},
where several agents transport one object to the destination via pushing or grasping.
These motion planning problems remain challenging due to the dynamic and geometric constraints
associated with different tasks.
Moreover, these works mostly assume a global objective for the whole team,
i.e., without the need for decomposing and assigning subtasks.

% STAMP
Integrating the above two aspects yields a task and motion planning (TAMP) problem
for multi-agent systems~\cite{guo2016communication, guo2018multirobot}.
The work in~\cite{hartmann2020robust} proposes a combination of receding horizon-based task decomposition
and motion planning for autonomous assembly.
Similarly, a furniture-assembly task is considered in~\cite{dogar2019multi}
where re-grasps are introduced to decompose the long-horizon assembly operations.
Both works put strong emphases on the physical stability and sequential feasibility
during collaborative manipulation,
while neglecting the combinatorial aspect as only a few agents are considered.
The classic game of ``cops and robbers'' is considered
in~\cite{pierson2016intercepting, chen2016multiplayer}
where multiple pursuers are tasked to capture multiple evader.
The proposed solutions decouple the task assignment and the controller design
by adopting a greedy assignment policy, e.g., nearest target~\cite{pierson2016intercepting}
or maximum matching pairs~\cite{chen2016multiplayer}.
Thus, there remains a need for an integrated solution
for multi-agent TAMP problems that can tackle  simultaneously
the combinatorial task assignment and the collaborative control design.

%==============================
\subsection{Our Method}\label{subsec:intro-our}
To overcome these challenges, this work proposes a combinatorial-hybrid optimization (CHO)
framework for multi-agent systems under collaborative tasks.
Several agents can collaborate on one common task as a coalition,
the cost of which
depends not only on the number of participants, but also the mode of collaboration
and the underlying control parameters.
Thus, the proposed framework includes two layers:
(i) the optimization of coalition structure,
and (ii) the hybrid search of optimal collaborative behavior
as the sequence of discrete modes and continuous control parameters.
Both layers are performed concurrently and in-demand, i.e.,
the hybrid search for a given task coalition is solved for the actual cost,
only when such a coalition is demanded promising during task assignment.
It is shown that the final solution of coalition structure and
the associated collaborative behavior
is Nash-stable.
To demonstrate its applicability,
two non-trivial multi-agent collaborative tasks,
including collaborative transportation and dynamic capture,
are modeled and solved via the proposed framework.

Main contribution of this work lies in two aspects:
(i) the formulation of combinatorial-hybrid optimization problems for multi-agent systems
under collaborative tasks, particularly for scenarios
where the agent-per-task costs do not exist and can only be
derived after solving a hybrid optimization problem;
(ii) the proposed framework finds simultaneously the coalition formation
and the optimal collaborative behaviors, with a provable quality guarantee.

%%========================================
\section{Problem Description}\label{sec:problem}

%==============================
\subsection{Model of Workspace and Agents}\label{subsec:ws}
Consider a team of~$N$ agents that collaborate in a shared workspace,
denoted by~$\mathcal{X}\subset \mathbb{R}^X$.
The system state~$x\in \mathcal{X}$ includes not only the agent states
but also other dynamic components such as movable objects and targets.
Due to the dynamic and geometric constraints such as collision avoidance among the agents
and with obstacles,
the system is required to stay within the allowed subset~$\mathcal{X}_{\texttt{safe}}\subset \mathcal{X}$.

%==============================
\subsection{Parameterized Modes}\label{subsec:modes}
Moreover, these agents can change the system state by numerous \emph{parameterized modes},
denoted by~$\Xi\triangleq \{\xi_1,\cdots, \xi_K\}$.
Under each mode~$\xi_k\in \Xi$,
the system state evolves under a closed-loop dynamics
$h_k:\mathcal{X} \times 2^{\mathcal{N}} \times \mathbb{R}^{P_k} \rightarrow \mathcal{X}$,
i.e.,
\begin{equation}\label{eq:mode}
x(t+1) \triangleq h_k\big{(}x(t),\,\mathcal{N}_k,\, \rho_k\big{)},\; \forall t \in [t_0,\, t_0+T_k],
\end{equation}
where~$k\in \mathcal{K}\triangleq\{1,\cdots,K\}$ for one valid mode;
$\mathcal{N}_k\subseteq \mathcal{N}$ is a subset of agents that participate in this mode
(called coalitions);
$\rho_k\in \mathbb{R}^{P_k}$ is the continuous parameter chosen for this mode with dimension~$P_k$;
$x(t)$ and $x(t+1)$ are the system states before and after agents in~$\mathcal{N}_k$
perform mode~$\xi_k$ with parameter~$\rho_k$ for one time step;
$t_0\geq 0$ is an arbitrary starting time;
$T_k$ is a given minimum duration of mode~$\xi_k$.
For performance measure, there is a cost function~$c_k:\mathcal{X} \times 2^{\mathcal{N}} \times \mathbb{R}^{P_k} \rightarrow \mathbb{R}^+$
associated with each mode~$\xi_k$ under a particular choice of~$(\mathcal{N}_k,\,\rho_k)$.
It is assumed in this work that the functions~$h_k,\,c_k$
above associated with each mode~$\xi_k\in \Xi$
is {accessible} either via explicit functions or numerical simulations.
Such modes are often built upon well-established functional modules that are designed
beforehand for specific and simpler purposes.
More examples for two different applications can be found in Sec.~\ref{sec:cases}.

%==============================
\subsection{Collaborative Tasks}\label{subsec:tasks}
Furthermore, there are~$M$ tasks specified for the team,
denoted by~$\Omega\triangleq \{\omega_1,\cdots,\omega_M\}$.
Each task~$\omega_m\in \Omega$ in the most general sense is to change
the system state to a set of goal states~$\mathcal{X}_{G_m}\in \mathcal{X}$.
Each task can be accomplished by a \emph{hybrid plan}
as a sequence of modes with appropriate choices of coalitions and parameters, i.e.,
\begin{equation}\label{eq:task-accomp}
  \varphi_m \triangleq (\xi_{k^m_1},\, \mathcal{N}_{k^m_1},\, \rho_{k^m_1}) \cdots
  (\xi_{k^m_L},\, \mathcal{N}_{k^m_L},\, \rho_{k^m_L}),
\end{equation}
where $L>0$ is the total number of modes to be optimized;
and $(\mathcal{N}_{k^m_1},\, \rho_{k^m_1})$ are permissible coalitions and parameters
according to function~$h_{\xi_k}$ in~\eqref{eq:mode}
for each mode $\xi_{k^m_\ell}\in \Xi$, $\forall \ell=1,\cdots, L$.
Thus, the evolution of system state under plan~$\varphi_m$ is constrained by:
\begin{equation}\label{eq:task-state}
\begin{split}
  x_{k^m_\ell}&= h_{k^m_\ell}\big{(}x_{k^m_{\ell-1}},\, \mathcal{N}_{k^m_\ell},\, \rho_{k^m_\ell}\big{)};\\
  x_{k^m_0} & = x_0,\; x_{k^m_L} \in  \mathcal{X}_{G_m},
\end{split}
\end{equation}
where~$x_0\in \mathcal{X}$ is a given initial state.
The associated cost of~$\varphi_m$ is given by the accumulated cost, i.e.,
$c(\varphi_m)\triangleq \sum_{\ell} c_{k^m_\ell}(x_{k^m_{\ell-1}},\, \mathcal{N}_{k^m_\ell},\, \rho_{k^m_\ell}),$
which holds for each~$\omega_m\in \Omega$.
In addition, since different modes can be executed in a concurrent way
for different tasks, it is assumed in this work that
different tasks change different dimensions of the state in an independent way,
i.e.,
\begin{equation}\label{eq:compose-task}
  x(t+1) = x(t) + \sum_{\omega_m\in \Omega}  \Big{(}h_{k^m_t}\big{(}x(t),\,
  \mathcal{N}_{k^m_t},\, \rho_{k^m_t}\big{)}-x(t)\Big{)},
\end{equation}
where~$\xi_{k^m_t}\in \Xi$ is the active mode of task~$\omega_m$ at time~$t\geq 0$;
$(\mathcal{N}_{k^m_t},\,\rho_{k^m_t})$ is the associated coalition and parameter;
$x(t)$ is the current system state;
and $x(t+1)$ is the resulting state after executing all active modes for one time step.
Lastly, since each agent can only participate in maximal one task for all time,
it holds that:
\begin{equation}\label{eq:non-common}
    \mathcal{N}_{m_1}(t) \cap \mathcal{N}_{m_2}(t) =\emptyset,
    \;\forall \omega_{m_1},\omega_{m_2}\in \Omega;
\end{equation}
where~$\mathcal{N}_{m_1}(t), \mathcal{N}_{m_2}(t)$ denote the coalitions
responsible for executing any two
tasks~$\omega_{m_1},\omega_{m_2}\in \Omega$ at time~$t\geq 0$.

%==============================
\subsection{Combinatorial-hybrid Optimization (CHO)}\label{subsec:prob}
\begin{problem}\label{problem:overall}
Given the above model of~$N$ agents and~$M$ tasks,
the complete combinatorial-hybrid optimization (CHO) problem is defined as follows:
\begin{equation}\label{eq:cho}
  \begin{split}
    \underset{\{\varphi_m\}}{\textbf{min}}&\quad \left\{ \underset{m}{\textbf{max}}\,
    \big{\{}c(\varphi_m)\big{\}} +\frac{1}{M}\, \sum_m c(\varphi_m) \right\}\\
    \textbf{s.t.}
    &\quad \eqref{eq:task-state}-\eqref{eq:non-common},\, \forall \ell,\forall m;
\end{split}
\end{equation}
where $\{\varphi_m\}$ is the set of hybrid plans for all tasks;
the objective is to minimize a balanced cost between the maximum cost and the average cost among all tasks;
\eqref{eq:task-state}-\eqref{eq:non-common} are the dynamic and geometric
constraints associated with the system state and the structure of coalitions.
\hfill $\blacksquare$
\end{problem}

%% In other words, each agent should decide at each time step on the its task coalition,
%% and the associated mode and parameter.
%% Thus, the complexity of the above problem is combinatorial
%% w.r.t. not only the number of agents and tasks,
%% but also the length of task plan and the number of modes.

%%========================================
\section{Two Concrete Use Cases}\label{sec:cases}
To clarify the problem formulation described in Sec.~\ref{sec:problem},
two use cases of multi-agent systems
that can be modeled by the CHO framework are presented first in this section,
of which the detail setup and results are given in Sec.~\ref{sec:experiments}.

%==============================
\begin{figure}[t]
  \centering
  \includegraphics[width=0.9\linewidth]{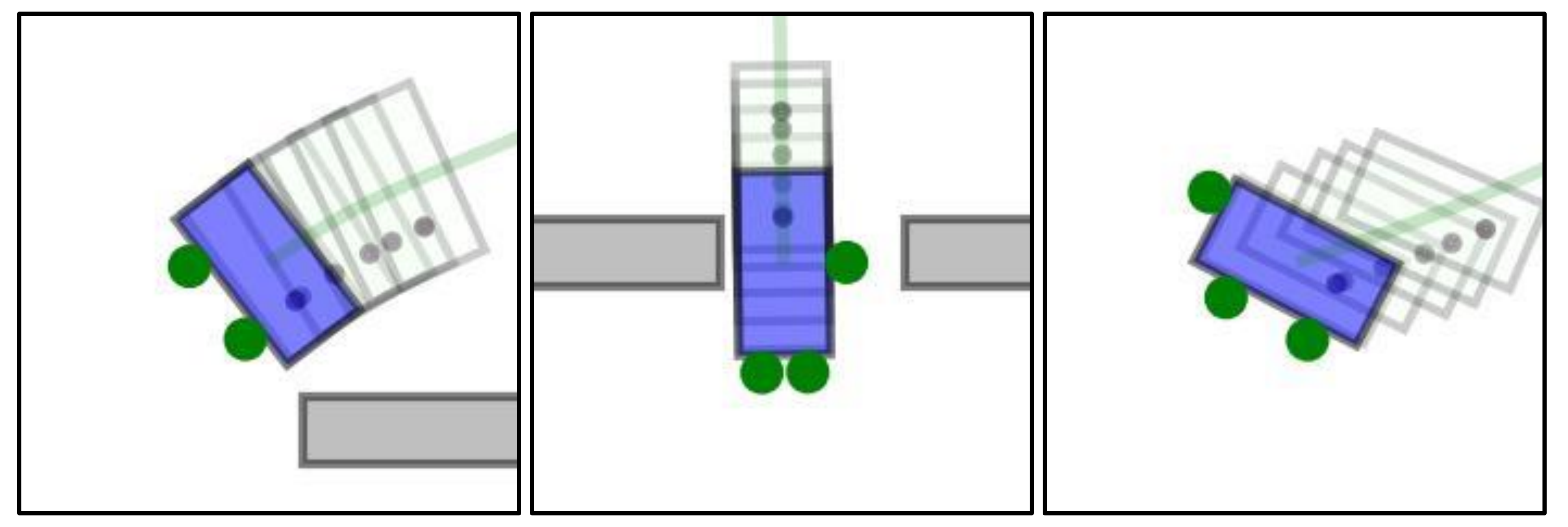}
  \caption{Illustration of three different modes for the use case of object transportation
    described in Sec.~\ref{subsec:col-transport}:
    Long-Side Pushing (\textbf{Left}), Short-Side Pushing (\textbf{Middle}),
    and  Diagonal Pushing (\textbf{Right}).
  }
  \label{fig:transportation_mode}
  \vspace{-2mm}
\end{figure}
%==============================
%==============
\subsection{Collaborative Transportation}\label{subsec:col-transport}
Consider the problem in which a team of~$N$ agents is tasked to move~$M$
identical rectangular boxes
from their initial positions to target positions in a cluttered workspace,
as illustrated in Fig.~\ref{fig:transportation_mode}.
The state of a box is determined by its center~$(x_m, y_m)$ and the its orientation~$\psi_m$,
of which the target position is given by~$(x_m^\star,y_m^\star)$.

The agents can make contact at specific points on the box and push it forward
by applying pushing forces $F_n\in[0,\, F_{\max}]$.
Different combinations of contact points result in different system dynamics thus different modes~$\Xi=\{\xi_k\}$.
As shown in Fig.~\ref{fig:transportation_mode}, the box can be pushed by different number of agents in different modes,
where the parameters are the applied forces~$\rho_k=\{F_n\}$.
Clearly, more agents lead to a higher degree of controllability
and faster motion.
Moreover, a sequence of different modes with distinct parameters might be required
to move a box to its target position, e.g., through narrow passages and sharp corners.

%==============================
\begin{figure}[t]
  \centering
  \includegraphics[width=0.9\linewidth]{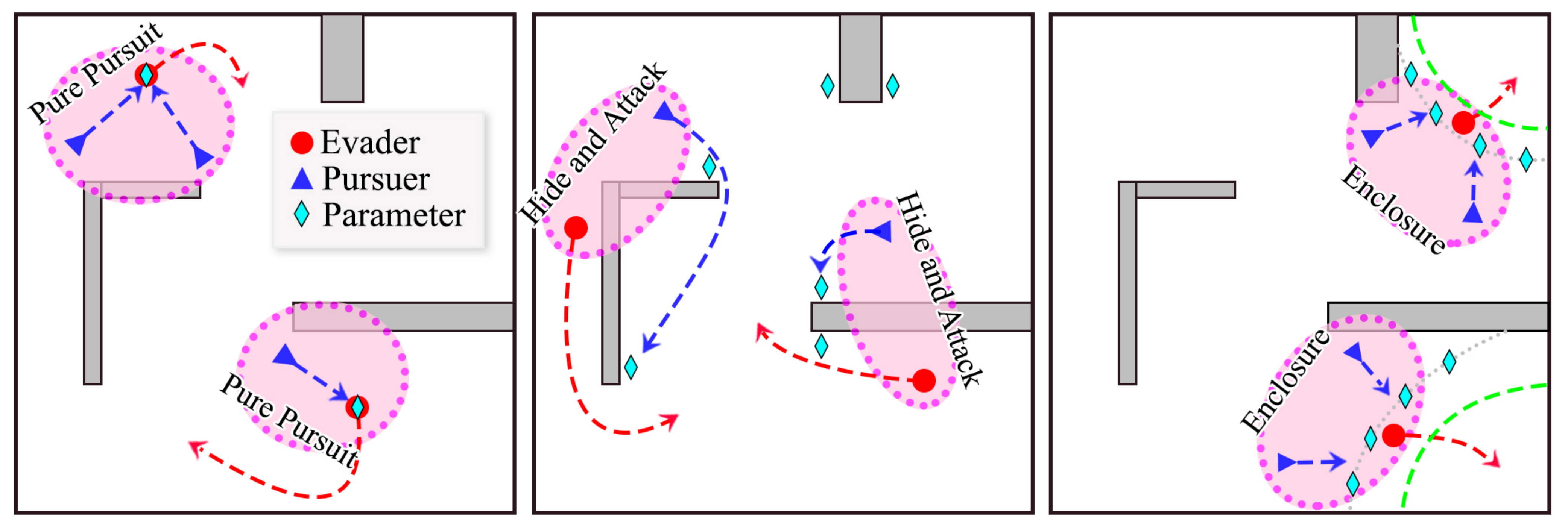}
  \caption{Illustration of three different modes associated with the use case of dynamic capture
    described in Sec.~\ref{subsec:dynamic-capture}:
    Pure Pursuit (\textbf{Left}), Hide and Attack (\textbf{Middle}),
    and Enclosure (\textbf{Right}).}
  \vspace{-2mm}
  \label{fig:capture_mode}
\end{figure}
%==============================

%==============
\subsection{Dynamic Capture}\label{subsec:dynamic-capture}
Consider the problem of dynamic capture
that involves~$N$ pursuers and~$M$ evaders in a cluttered 2D workspace.
Their states and velocities are denoted by~$(x_n(t), v_n(t))$ and $(y_m(t), v_m(t))$.
The agents all have the \emph{same} maximum velocity~$v_{\max}$,
meaning that collaborations are required for the pursuers to catch the evaders.
There is a capture task associated with each evader,
which moves in the opposite direction of all pursers, weighted their relative distances.

In addition, as shown in Fig.~\ref{fig:capture_mode},
there are three parameterized modes~$\Xi=\{\xi_1,\,\xi_2,\, \xi_3\}$ for each task:
(i) \emph{Pure Pursuit}.
The pursuers follow the ``line-of-sight'' strategy to move towards the assigned evader,
i.e., $v_n=v_{\max}\, \texttt{point}(\{x_n\},\,\rho_1)$ as the unit vector from~$x_n$
to the parameter~$\rho_1\in \mathbb{R}^2$ determined by the position of assigned evader.
(ii) \emph{Hide and Attack}.
The pursuers conceal themselves behind obstacles until the assigned evader appears,
and then encircle it,
i.e., $v_n = \texttt{nav}(\{x_n\},\, \rho_2)$ as a navigation function that guides
the pursuer to a hiding or attacking position~$\rho_2\in \mathbb{R}^2$.
(iii) \emph{Enclosure}.
The pursers drive the assigned evader into a corner by reducing the region of
advantage~\cite{8453470} for the evader until it is captured,
i.e., $v_n = \texttt{corner}(\{x_n\},\,\rho_3)$
as an optimization algorithm that generates the optimal velocity
given other pursuers~$\{x_n\}$ and key position~$\rho_3\in \mathbb{R}^2$.
Consequently, each pursuer should choose a sequence of modes,
coalitions and the associated parameters, to capture all evaders as fast as possible.

%%========================================
\section{Proposed Solution}\label{sec:solution}
The proposed solution tackles Problem~\ref{problem:overall} via two interleaved and concurrent layers:
(i) coalition formation, and (ii) hybrid optimization of collaborative behaviors.
The first layer proposes candidates of coalitions for each task to minimize the overall cost.
These candidates are then sent to the second layer which solves a
constrained hybrid optimization for each coalition,
to determine the optimal hybrid plans and actual cost to accomplish the assigned task.
These costs are then fedback to the first layer to adjust existing coalitions.
This process repeats itself until a Nash-stable solution is found.
Each layer is described in detail in this section.

%%========================================
\subsection{Coalition Formation}\label{subsec:coalition}
\subsubsection{Problem of Coalition Formation}
The first layer assigns the set of tasks to available agents.
To begin with, the task assignment problem is formulated
based on the literature of coalition formation~\cite{rahwan2015coalition,li2020anytime}.
Particularly, consider the coalition structure defined as follows:
\begin{equation}\label{eq:general-form}
 \mathcal{F} \triangleq (\mathcal{R},\,\Omega,\, f),
\end{equation}
where~$\mathcal{R}=\{1,\cdots,N\}$ is the team of~$N$ agents;
$\Omega=\{\omega_1,\cdots, \omega_M\}$ is the set of~$M$ tasks;
$f:2^{\mathcal{R}} \times \Omega \rightarrow \mathbb{R}^+$ is the cost function
of a potential coalition for any given task in~$\Omega$.
Not that since the exact cost~$f$ above is often unknown,
thus estimated initially by simple heuristics such as Euclidean distance, denoted by~$\widehat{f}$.
%===========================
\begin{figure}[t]
  \centering
  \includegraphics[width=0.9\linewidth]{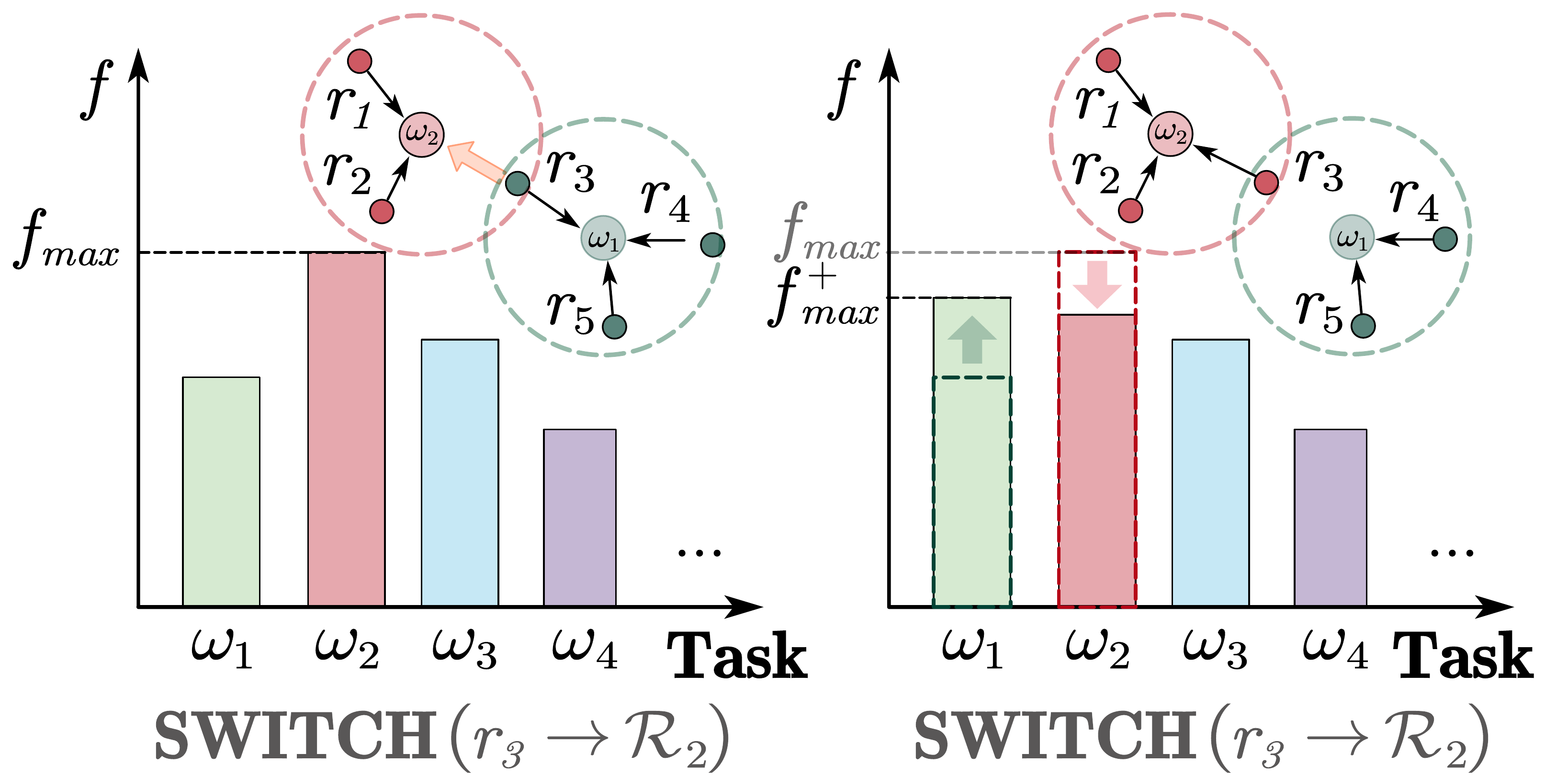}
  \vspace{-7mm}
  \caption{Illustration of the coalition formation algorithm
    in Sec.~\ref{subsec:task-assignment}.
    Agents switch coalitions to reduce the overall cost in~\eqref{eq:all-cost}.}
  \label{fig:assignment}
  \vspace{-2mm}
\end{figure}
%===========================
%====================
\begin{definition}[Assignment]\label{def:assignment}
A valid solution of the coalition structure~$\mathcal{F}$ in~\eqref{eq:general-form} is called an assignment,
denoted by~$\mu =\{(\mathcal{R}_m,\,\omega_m),\,\forall \omega_m \in \Omega\}$,
where~$\mathcal{R}_m \subset \mathcal{R}$ is a coalition
that performs the common task~$\omega_m \in \Omega$,
and~$\mathcal{R}_{m_1}\cap\mathcal{R}_{m_2}=\emptyset,\, \forall m_1\neq m_2$.
Moreover, the estimated cost is given by:
\begin{equation}\label{eq:all-cost}
  C(\mu) \triangleq \underset{\omega_m\in \Omega}{\textbf{max}}\, \{\widehat{f}_m\}
  + \frac{1}{M}\sum_{\omega_m\in \Omega}\, \widehat{f}_m,
\end{equation}
where~$\widehat{f}_m \triangleq \widehat{f}(\mathcal{R}_m,\, \omega_m)$ is the estimated cost of each coalition;
and $C(\cdot)$ is a balanced cost based on~\eqref{eq:cho}.
\hfill $\blacksquare$
\end{definition}
%====================

Let $\mu(\omega_m)=\mathcal{R}_m$ return the coalition for task~$\omega_m$
and $\mu(n)=\omega_m$ return the  task assigned to agent~$n$,
$\forall n \in \mathcal{R}_m$.
More importantly, an assignment~$\mu$ can be modified
via the following {switch} operation.
%====================
\begin{definition}[Switch Operation]\label{def:switch}
  The operation that agent~$n \in\mathcal{R}$ is assigned to
  task~$\omega_{m} \in \Omega$ is called a \emph{switch} operation,
  denoted by~$\sigma^{m}_{n}$.
  The switch~$\sigma^{m}_{n}$ is valid only if agent~$n$ can perform task~$\omega_m$.
  Thus, an assignment~$\mu$ is changed into a new assignment~$\hat{\mu}$
  via~$\sigma^{m}_{n}$ such that~$\hat{\mu}(n)=\omega_m$.
  For brevity, denote by~$\hat{\mu}=\sigma^{m}_{n}(\mu)$.
  \hfill $\blacksquare$
\end{definition}
%====================

%====================
\begin{definition}[Nash Stable]\label{def:nash}
  An assignment~$\mu^\star$ is called Nash-stable,
  if there does not exist any switch operation~$\sigma^m_n$ such that~$C(\sigma^{m}_{n}(\mu))<C(\mu^\star)$ by~\eqref{eq:all-cost},
  $\forall n\in \mathcal{N}$.
  \hfill $\blacksquare$
\end{definition}
%====================

Given the above definitions, the first layer of coalition formation is transformed into the problem of
finding a Nash-stable assignment as follows.

%====================
\begin{problem}\label{prob:nash}
  Given a coalition structure~$\mathcal{F}$ in~\eqref{eq:general-form},
  find one Nash-stable assignment~$\mu^\star$ by Def.~\ref{def:nash}.
  \hfill $\blacksquare$
\end{problem}
%====================

%====================
\subsubsection{Dynamic Task Assignment}\label{subsec:task-assignment}
The proposed solution follows an iterative switching operations to reduce
the largest cost of all tasks.
An initial assignment~$\mu_0$ can be derived in a random or greedy way.
Then, the \emph{target} coalition with the $k$-th maximum cost is selected via:
\begin{equation}\label{eq:target-coal}
  (\mathcal{R}_{m^\star}, \omega_{m^\star}) = \textbf{argmax}^{[k]}_{(\mathcal{R}_m,\,\omega_m)\in \mu_0}\, \{\widehat{f}_m\},
\end{equation}
where $k=1$ initially for the maximum cost;
and ties are broken arbitrarily.
Given~$(\mathcal{R}_{m^\star}, \omega_{m^\star})$, the first step is to calculate the \emph{actual} utility~$f_{m^\star}$,
via the hybrid optimization procedure described in the subsequent section.
There are two possible outcomes:
(I) If $f_{m^\star}\geq \textbf{max}_{m\neq m^\star}\{\widehat{f}_m\}$,
it means that~$\omega_{m^\star}$ remains the target coalition.
As illustrated in Fig.~\ref{fig:assignment}, the goal is to find a switch operation~$\sigma^{m^\star}_n$ for one agent~$n\in \mathcal{N}$
such that after applying~$\sigma^{m^\star}_n$ the maximum cost is reduced, i.e.,
\begin{equation}\label{eq:condition}
  \begin{split}
  &\textbf{max}\,\left\{{f}\big{(}\mathcal{R}_{m^\star} \cap \{n\},\, \omega_{m^\star}\big{)},\,
    {f}\big{(}\mathcal{R}_{m^-}\backslash \{n\},\, \omega_{m^-}\big{)} \right\} \\
    &<
  \textbf{max}\,\left\{\widehat{f}(\mathcal{R}_{m^\star},\, \omega_{m^\star}),\,
  \widehat{f}(\mathcal{R}_{m^-},\, \omega_{m^-})\right\},
\end{split}
\end{equation}
where~$m^-=\mu_0(n)$ is the task to which agent~$n$ was assigned \emph{before} applying the switch.
Such a switch can be found by iterating through all agents within~$\mathcal{R}\backslash\mathcal{R}_{m^\star}$,
and verifying whether the condition in~\eqref{eq:condition} holds.
Once a switch~$\sigma^{m^\star}_n$ is found, a new assignment is given by~$\mu_1=\sigma^{m^\star}_n(\mu_0)$,
after which a new target coalition can be found via~\eqref{eq:target-coal}.
(II) If $f_{m^\star}<\textbf{max}_{m\neq m^\star}\{\widehat{f}_m\}$,
it means that~$\omega_{m^\star}$ is no longer the target coalition based on its actual cost.
Then, the estimated utility $\widehat{f}_{m^\star}$ is updated accordingly
and thus a new target coalition can be found via~\eqref{eq:target-coal}.
The above procedure is repeated until the target coalition does not change anymore,
of which the assignment and target coalition are denoted by~$\mu_{K_1}$
and~$(\omega_{m_1^\star}, \mathcal{R}_{m_1^\star})$, respectively.

Afterwards, the same process is repeated but focusing on the target coalition with \emph{second} largest cost,
i.e., by setting~$\mu_{K_1}$ as the initial assignment and
$k=2$ instead in~\eqref{eq:target-coal}.
Similarly, the procedure converges in this round after the target coalition remains unchanged.
The same round is performed for~$k=3$ and so on until~$k=M$, when
the target coalition~$\mu_{K_M}$ has the minimum utility.
Note that if the target coalitions from the previous rounds, e.g., round~$k_1$, are changed during the current round~$k_2>k_1$, the whole process is re-started from~$k=1$.

%====================
\begin{proposition} \label{th:local_optimal}
  The final assignment~$\mu_{K_M}$ is a Nash-stable assignment of~$\mathcal{F}$
  under the actual cost function~$f$.
\end{proposition}
\begin{proof}
  (Sketch) First, the costs of all coalitions in~$\mu_{K_M}$ are
  the actual cost derived by hybrid optimization.
  Second, the fact that no switch operations can be found for each agent fulfills the Nash-stable condition in Def.~\ref{def:nash}.
\end{proof}

%%========================================

%%========================================
\subsection{Hybrid Optimization}\label{subsec:hybrid}

% TODO: HybridSearchNode: X,M,A
As previously mentioned,
the group of agents in~$\mathcal{R}_m$ as a coalition should follow
a hybrid plan~$\varphi_m$ as defined in~\eqref{eq:task-accomp} to accomplish task~$\omega_m$.
This section presents how such a hybrid can be found via the proposed hybrid optimization.

\subsubsection{Problem of Hybrid Optimization}
For the ease of notation,
let $\mathbf{X}=x_{k_0^m} \cdots x_{k_T^m}$
be the sequence of system states at discrete time
steps~$t=0,1,\cdots,T$ for a sufficient duration~$T$;
$\mathbf{\Xi}=\xi_{k_{t_1}^m},\cdots,\xi_{k_{t_N}^m}$
and $\mathbf{P}=\rho_{k_{t_1}^m},\cdots,\rho_{k_{t_N}^m}$
be the sequence of modes
applied to the system for time periods
$[t_0,t_1),\, [t_1,t_2),\, \cdots,\, [t_{N-1},t_N)$,
where $t_n=nT_0, T_0 \in \mathbb{N}$, $\forall n=0,\cdots,\lfloor \frac{T}{T_0}\rfloor$.
% \todo{$n$ is also used for indexing agents?}
Note that $T_0>0$ is chosen as a lower bound on the duration of each mode
to avoid too frequent switching of modes and parameters.
Furthermore,
let $\mathbf{X}_t,\,\mathbf{\Xi}_t,\,\mathbf{P}_t$ denote $x_{k_t^m},\, \xi_{k_t^m},\, \rho_{k_t^m}$.
Then, the hybrid optimization problem is stated as follows:
\begin{problem}\label{problem-ho}
  Given a task $\omega_m$ and the associated coalition $\mathcal{R}_m$,
  find the optimal sequences~$(\mathbf{X},\, \mathbf{\Xi},\, \mathbf{P})$
  that solve the hybrid optimization (HO) problem below:
\begin{equation}\label{eq:HybridOpt}
    \begin{split}
 &\quad\quad\quad\underset{\mathbf{\Xi},\mathbf{P}}{\textbf{min}}\, \sum^T_{t=0} c_{\textup{cont}}(\mathbf{X}_t,\, \mathbf{P}_{t})\\
 \textbf{s.t.} &\quad \mathbf{X}_0=x_0,\; \mathbf{X}_T\in \mathcal{X}_{G_m};\\
 &\quad \mathbf{\xi}_{k_t^m}=\mathbf{\Xi}_{t_n},\; \rho_{k_t^m}=\mathbf{P}_{t_n},\; \forall t \in [t_n,t_{n+1});\\
 &\quad \mathbf{X}_{t+1}= h_{\xi_{k_t^m}}(\mathbf{X}_{t},\, \mathcal{R}_m,\, \mathbf{P}_{t}),\; \forall t\in [0,\, T];
    \end{split}
\end{equation}
where~$c_{\textup{cont}}(\cdot)$ is a general function including the control cost and smoothness;
$h_{\xi_{k_t^m}}(\cdot)$ is the system dynamics under each mode from~\eqref{eq:mode};
and the constraints require that the mode and parameter are kept static within~$[t_n,\,t_{n+1})$.
\hfill $\blacksquare$
\end{problem}

Note that different from the original combinatorial-hybrid optimization problem in~\eqref{eq:cho},
the goal of the hybrid optimization problem above is to find the optimal hybrid plan $\varphi_m$ for a specific task $\omega_m$
and the corresponding coalition $\mathcal{R}_m$.
A novel hybrid-search algorithm called Heuristic Gradient-Guided Hybrid Search (HGG-HS)
is proposed to solve Problem~\ref{problem-ho}.
Instead of feeding the above problem directly into a general nonlinear optimizer,
the proposed algorithm combines: (i) the $A^\star$-based discrete search
for the optimal sequence of modes,
and (ii) the gradient-based optimization for the optimal sequence of parameters.
More specifically, the hybrid search tree is defined as~$\mathcal{T}\triangleq (V,\,E,\,\nu_0,\,V_G)$,
where $V=\{\nu\}$ is a set of vertices for~$\nu\in \mathcal{X}_{\texttt{safe}}$;
$E\subset V\times V$ is a set of edges; $\nu_0\in V$ is the initial vertex;
and $V_G \subset V$ is a set of goal states determined by~$\mathcal{X}_{G_m}$.
Moreover, let $cost(\nu)$ be the cost of node $\nu$
and $prev(\nu)$ be the parent node of $\nu$.

\subsubsection{Design of Heuristic Functions}
As stated in~\cite{cui2020heuristic},
a proper design of the heuristic function~$h: \mathcal{X} \rightarrow \mathbb{R}^+$
is essential for the performance of the $A^\star$ search algorithm.
Since it is impractical to find an exact heuristic function~$h^{\texttt{opt}}$
that estimates \emph{perfectly} the cost from a given vertex~$\nu\in V$ to the goal set~$V_G$,
this work proposes two approximations of the exact heuristics at different level of abstraction:
(i) the global approximation $h^{\mathcal{G}}$ that serves as a lower bound on the actual cost,
i.e., $h^{\texttt{opt}}(x) \geq h^{\mathcal{G}}(x)$, $\forall x\in \mathcal{X}$.
For instance, Euclidean distance is a common choice as admissible heuristics;
(ii) the local approximation $h^{\mathcal{L}}$ that is differentiable
and has similar gradients to~$h^{\texttt{opt}}$ in a local neighborhood, i.e.,
$\|\nabla h^{\mathcal{L}}_{x_0}(x) - \nabla h^{\texttt{opt}}(x)\|\leq E$,
$\forall x\in \mathrm{U}_d(x_0)=\{x\in\mathcal{X}|\| x-x_0 \|\leq d\}$ and $E>0$.
Furthermore, a balanced heuristic function $h^\mathcal{B}$ is defined recursively as follows:
\begin{equation}\label{eq:balanced}
  h^\mathcal{B}(\nu)=\lambda \Big{(}h^\mathcal{B}(\tilde{\nu})+\Delta h^{\mathcal{L}}(\tilde{\nu},\,\nu)\Big{)}
  +(1-\lambda)\, h^\mathcal{G}(\nu),
\end{equation}
where~$\tilde{\nu}=prev(\nu)$;
$\lambda \in [0,1]$ is a weighting factor;
and $\Delta h(\tilde{\nu},\,\nu)$ is the change of cost from~$\tilde{\nu}$ to~$\nu$,
which is estimated by the accumulated change along a path, i.e.,
\vspace{-1mm}
\begin{equation}\label{eq:cost-change}
    \Delta h^{\mathcal{L}}(\tilde{\nu},\, \nu)=\sum_{\ell=1}^{L}\Big{(}h_{x_\ell}^{\mathcal{L}}(x_\ell)
    -h_{x_\ell}^{\mathcal{L}}(x_{\ell-1})\Big{)},
\end{equation}
where $x_{0}=\tilde{\nu}$, $x_{L}=\nu$,
and $x_{\ell}\in\mathrm{U}_d(x_{\ell-1})$, $\forall \ell=1,\cdots,L$.
Moreover, $\lambda$ is a parameter that effects the greedy-ness of HGG-HS.
Namely, when $\lambda=0$ holds, $h^{\mathcal{B}}$ reduces to~$h^{\mathcal{G}}$,
yielding a general heuristic search algorithm like~$A^\star$.
On the other hand, when $\lambda=1$ holds,
$h^{\mathcal{B}}$ only depends on the local approximation~$h^{\mathcal{L}}$,
resulting in a local greedy search.
%===========================
\begin{figure}[t]
  \centering
  \includegraphics[width=0.9\linewidth]{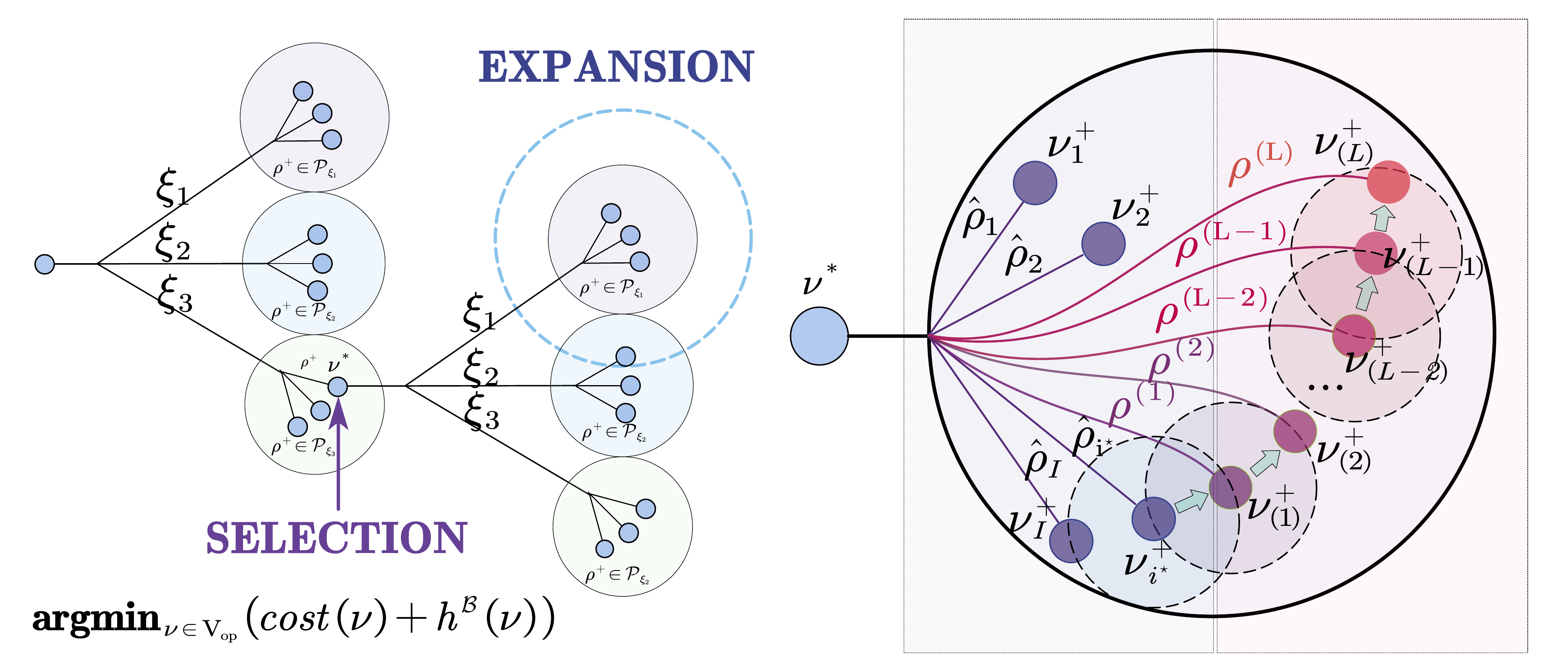}
  \caption{Illustration of the hybrid search algorithm in
        Sec.~\ref{sec:hgg-hs}:
        selection and expansion (\textbf{Left});
      iterative parameter optimization (\textbf{Right}).}
  \label{fig:HybridSearch}
  \vspace{-2mm}
\end{figure}
%===========================
\subsubsection{Heuristic Gradient-Guided Hybrid Search}
\label{sec:hgg-hs}
Given the above definition of balanced heuristics,
the hybrid search tree~$\mathcal{T}$ is explored via
an iterative process of node selection and expansion.
To begin with, similar to the $A^\star$ algorithm,
a priority queue $V_{\texttt{op}}\subset V$ is used to
store the vertices to be visited,
while a set $V_{\texttt{cl}}\subset V$ is used to store the vertices that have been fully explored.
Then, as illustrated in Fig.~\ref{fig:HybridSearch},
the proposed hybrid search algorithm consists of the following stages:
(I) \textbf{Selection}.
The vertex with the lowest estimated cost in~$V_{\texttt{op}}$ is selected,
i.e.,~$\nu^\star= \textbf{argmin}_{\nu\in V_{\texttt{op}}} \{cost(\nu)+h^{\mathcal{B}}(\nu)\}$,
of which the associated state is~$x^\star$.
(II) \textbf{Expansion}.
This vertex~$\nu^\star$ is expanded in three steps:
(i) first, a feasible mode~$\xi \in\Xi$ is chosen given the
state~$x^\star$;
(ii) then, a set of candidate parameters~$\{\rho^\star\} \subset \mathcal{P}_{\xi}$ is found for
mode~$\xi$ and~$x^\star$ via an iterative optimization in the parameter space
as described in the sequel;
(iii) lastly, given~$\xi$ and~$\{\rho^\star\}$ above,
the set of resulting child vertices~$\mathcal{C}_\xi(\nu^\star)\triangleq \{\nu^+\}$ is given by:
\begin{equation}\label{eq:expand}
  \begin{split}
    &\mathbf{x}_{t+1}=h_{\xi}(\mathbf{x}_t,\,\mathcal{R}_m,\,\rho^\star),\; \forall t\in [0,\, T_0]; \\
    & {x}_0={x}^\star,\; \nu^+=\mathbf{x}_{T_0};
    \end{split}
\end{equation}
which is encapsulated by~$\nu^+\triangleq \texttt{Expand}(x^\star,\,\rho^\star)$
for the ease of notation.
Moreover, the properties of $\nu^+$ is updated by
$cost(\nu^+) = cost(\nu^\star)+\sum_{t=0}^{T_0}c_{\text{cont}}(\mathbf{x}_t,\,\rho^\star)$
and $prev(\nu^+) = \nu^\star$.
Each child node $\nu^+\in\mathcal{C}_\xi(\nu^\star)$ is added
to the node set~$V$ and~$V_{\texttt{op}}$,
if
$\nu \notin V_{\texttt{cl}}$ and $cost(\nu^+)\leq cost(\nu)$,
$\forall \nu\in V$ that satisfying $[(\nu^+-\nu)/\delta]=\mathbf{0}$,
where $[\cdot]$ is the rounding function for a radius~$\delta>0$.
Afterwards, the edge~$(\nu^\star,\nu^+)$ is added to the edge set~$E$
and labeled by the associated mode and parameter~$(\xi,\,\rho^\star)$.
(III) \textbf{Termination}.
If all the child vertices of~$\nu^\star$ have been explored,
$\nu^\star$ is removed from $V_{\texttt{op}}$ and added to $V_{\texttt{cl}}$.
More importantly,
if $\nu^\star\in V_G$ holds,
the optimized sequences $\mathbf{\Xi}$ and $\mathbf{P}$
as the solution to Problem~\ref{problem-ho} can be obtained by
tracing back the parent vertices and
retrieving its label~$(\xi,\,\rho)$.
Thus, the hybrid search algorithm returns the hybrid plan as described
in~\eqref{eq:HybridOpt},
along with the actual cost for the assigned task.

\subsubsection{Iterative Optimization of Mode Parameters}
The optimization for parameter~$\rho^\star$ in the stage of expansion above
for node~$\nu^\star$ follows a two-stage process.
In the first stage of \emph{primitive expansion},
parameters are initially selected from a predefined set of primitive parameters
$\{\hat{\rho}_1,\cdots,\hat{\rho}_I\}\subset \mathcal{P}_{\xi}$.
Then, a set of child vertices can be generated by $\nu_{i}^+=\texttt{Expand}(x^\star,\,\hat{\rho}_i)$,
$\forall \hat{\rho}_i\in\hat{\mathcal{P}}_{\xi}$.
Within this set, the child vertex with lowest estimated total cost is selected,
i.e.,
\begin{equation}\label{eq:estimate-cost}
\nu_{i^\star}^+=\textbf{argmin}_{\nu\in\{\nu^+_i\}}\Big{\{}cost(\nu)+h^{\mathcal{B}}(\nu)
+h^{\mathcal{L}}(\nu^\star,\,\nu)\Big{\}}
\end{equation}
and the associated parameter is~$\hat{\rho}_{i^\star}$.
In the second stage of \emph{iterative optimization},
the end state~$x_{\texttt{e}}^{(\ell)}$ and the associated parameter~$\rho^{(\ell)}$
is optimized via nonlinear optimization for iterations~$\ell=1,\cdots,L$.
Initially, $x_{\texttt{e}}^{(0)}=\nu_{i^\star}$ and~$\rho^{(0)}=\hat{\rho}_{i^\star}$.
Then, the following procedure is applied to update~$\rho^{(\ell)}$:
\begin{equation}\label{eq:udpate-rho}
  \begin{split}
    \rho^{(\ell+1)}&=\underset{\rho\in\mathcal{P}_{\xi}}{\textbf{argmin}}\,
    \Big{\{}\sum_{t=0}^{T_0}\, c_{\text{cont}}(\mathbf{x}_{t})+h_{x_{\texttt{e}}^{(\ell)}}^{\mathcal{L}}(\mathbf{x}_{T_0})\Big{\}};\\
    \textbf{s.t.} \quad &\mathbf{x}_{0}=\nu^\star,\; \mathbf{x}_{T_0}\in \mathrm{U}_d(x^{(\ell)}_{\texttt{e}})\; \text{in}~\eqref{eq:expand},
  \end{split}
\end{equation}
which can be solved by a general nonlinear optimization solver such as IPOPT~\cite{wachter2009short},
as all states are parameterized over~$\rho$.
Once~$\rho^{(\ell+1)}$ is obtained, the associated end state is updated by
$x^{(\ell+1)}_{\texttt{e}}=\texttt{Expand}(\nu^\star,\,\rho^{(\ell+1)})$.
This iterative process continues until iteration~$L$ such that
$\| x^{(L)}_{\texttt{e}}-x^{(L-1)}_{\texttt{e}}\|< d$.
Consequently, the set of pairs of parameters and end states is given
by~$\{(\rho^{(\ell)},\, x^{(\ell)}_{\texttt{e}})\}$,
of which the corresponding child vertices are~$\{\nu^+_{(\ell)}\}$.
Thus, the set of all child vertices $\mathcal{C}_{\xi}(\nu^\star)= \{\nu^+_{i},\forall i\}\cup\{\nu^+_{(\ell)},\forall \ell\}$ is sent to the next step in the stage of expansion.

\begin{theorem}
Consider a path~$\nu_0,\cdots,\nu_K$ of length~$K$ in~$\mathcal{T}$ obtained via the hybrid search algorithm.
Then, $h^{\mathcal{B}}_{k+1}\leq (1+\epsilon) h^{\texttt{opt}}_{k+1}$
holds, i.e., its cost is at most $(1 + \epsilon)$ times the minimum cost,
if $\lambda_k\leq\frac{h^{\mathcal{G}}_k}{EJ/\epsilon+\Delta c_m +h^{\mathcal{G}}_k}$
in~\eqref{eq:balanced}, $\forall k=1,\cdots,K$
and parameter $\epsilon\geq 0$.
\end{theorem}
\begin{proof}
For brevity,
let $h^{\mathcal{B}}_k,h^{\mathcal{G}}_k,h^{\texttt{opt}}_k$
denote the various heuristics of vertex $\nu_k$ defined in~\eqref{eq:balanced},
 $c_k=cost(\nu_k)$ and $\Delta h^\mathcal{L}_k=\Delta h^\mathcal{L}(\nu_k,\nu_{k+1})$.
To begin with, $h^{\mathcal{B}}_0=h^{\mathcal{G}}_0 \le(1+\epsilon) h^{\texttt{opt}}_0$
holds due to the boundary condition in~\cite{cui2020heuristic}.
Assuming that $h^{\mathcal{B}}_k\le(1+\epsilon) h^{\texttt{opt}}_k$ is satisfied for any~$k$.
Then, $\Delta h^{\mathcal{L}}_k$ can be bounded as follows:
\begin{equation*}
  \begin{split}
    \Delta h^{\mathcal{L}}_k&=\sum_{\ell=0}^{L-1}(h_{x_{\texttt{e}}^{(\ell)}}^{\mathcal{L}}(x^{(\ell)})-h_{x_{\texttt{e}}^{(\ell)}}^\mathcal{L}(x^{(\ell-1)}))\\
%%     &=\sum_{\ell=1}^{L-1}\int_{x_{\texttt{e}}^{(\ell)}}^{x^{(\ell+1)}}\nabla h^{\texttt{opt}}(x)\,\mathrm{d} x
%%     +\sum_{\ell=1}^{L-1}\int_{x_{\texttt{e}}^{(\ell)}}^{x^{(\ell+1)}}e(x)\,\mathrm{d} x\\
    &\leq h^{\texttt{opt}}_{k+1}-h^{\texttt{opt}}_k+EJ,
  \end{split}
\end{equation*}
where~$\| e(x) \| =\| \nabla h^{\mathcal{L}}_{x_0}(x) - \nabla h_{x_0}^{\texttt{opt}}(x)\| \leq E$;
$J$ is the maximum distance between a vertex and its parent vertex.
Then, it can be derived that:
\begin{equation}\label{eq:bound-h-B}
  \begin{split}
  h^{\mathcal{B}}_{k+1}&=\lambda(h^{\mathcal{B}}_k+\Delta h^{\mathcal{L}}_k)
  +(1-\lambda)(h^{\mathcal{G}}_{k+1})\\
  &\leq \lambda\big{(}(1+\epsilon) h^{\texttt{opt}}_k+\Delta h^{\mathcal{L}}_k\big{)}
  +(1-\lambda)\, h^{\mathcal{\texttt{opt}}}_{k+1}\\
  % &\leq \lambda((1+\epsilon) h^{\texttt{opt}}_k+h^{\texttt{opt}}_{k+1}-h^{\texttt{opt}}_k+EJd)\notag\\
  % &+(1-\lambda)h^{\mathcal{\texttt{opt}}}_{k+1}\notag\\
  %&\leq h^{\texttt{opt}}_{k+1}+\lambda\epsilon h^{\texttt{opt}}_k+\lambda EJd
  &\leq (1+\lambda\epsilon)h^{\texttt{opt}}_{k+1}+\lambda\epsilon\Delta c_m +\lambda EJ,
  \end{split}
\end{equation}
where $h^{\texttt{opt}}_k<h^{\texttt{opt}}_{k+1}+\Delta c$ and~$\Delta c=c_{k+1}-c_k <c_m$.
%% Then we have:
%% \begin{equation}
%%   \begin{split}
%%   h^{\mathcal{B}}_{k+1}&\leq(1+\lambda\epsilon)h^{\texttt{opt}}_{k+1}+\lambda\epsilon\Delta c_m +\lambda EJ
%%   \end{split}
%% \end{equation}
If $\lambda\leq\frac{h^{\mathcal{G}}_{k+1}}{EJd/\epsilon+\Delta c_m +h^{\mathcal{G}}_{k+1}}$ holds,
combining with~\eqref{eq:bound-h-B} implies $h^{\mathcal{B}}_{k+1}\leq (1+\epsilon) h^{\texttt{opt}}_{k+1}$.
Thus, via induction over~$k$, $h^{\mathcal{B}}_{K}\leq (1+\epsilon) h^{\texttt{opt}}_{K}$ holds,
which ensures a bounded sub-optimality.
\end{proof}
\subsection{Overall Framework}\label{subsec:overall}
As described in Sec.~\ref{subsec:task-assignment},
the two layers are executed in an interleaved and concurrent way.
More specifically, the layer of coalition formation searches for
the appropriate switch operations to reduce the overall cost in~\eqref{eq:all-cost}
of an assignment~$\mu$.
During this process, whenever the actual cost $f_{m}$ of a coalition $\mathcal{R}_m$
for task~$\omega_m$ is required,
the layer of hybrid optimization searches for the optimal sequence of modes and
parameters~$(\mathbf{\Xi}_m,\,\mathbf{P}_m)$ for the coalition $\mathcal{R}_m$
to accomplish~$\omega_m$, while minimizing the actual cost.
This process repeats itself until no such switch operations can be found
and the costs of all coalitions have been verified by the hybrid optimization.
The resulting assignment is given by~$\mu^\star$ and the optimal plan is
$(\mathbf{\Xi}_m^\star,\,\mathbf{P}_m^\star)$ for each task~$\omega_m\in \Omega$.
Thus, the final hybrid plan in~\eqref{eq:task-accomp} for~$\omega_m$
is given by:
\begin{equation}\label{eq:final-plan}
  \varphi_m^\star =
  (\xi^\star_{k^m_{t_1}},\, \mathcal{R}_m,\, \rho^\star_{k^m_{t_1}}) \cdots
    (\xi^\star_{k^m_{t_N}},\, \mathcal{R}_m,\, \rho^\star_{k^m_{t_N}}),
\end{equation}
where $\mathbf{\Xi}_m^\star=\xi^\star_{k^m_{t_1}}\cdots\xi^\star_{k^m_{t_N}}$ and
$\mathbf{P}_m^\star= \rho^\star_{k^m_{t_1}}\cdots\rho^\star_{k^m_{t_N}}$,
and the total length varies across different tasks.
Given these hybrid plans, each agent~$n\in \mathcal{N}$ can start executing
the assigned task~$\mu^\star(n)$, by following the optimal sequence of modes
with the chosen parameters.
Due to unforeseen disturbances and uncertainties,
the system state may evolve differently from the planned trajectory,
in which case the task assignment and hybrid plans should be updated
by resolving Problem~\ref{problem:overall} given the current system state.

%%========================================

%%========================================
%========================================
\section{Numerical Experiments} \label{sec:experiments}
To further validate the proposed method,
extensive numerical simulations are presented in this section.
The proposed method is implemented in Python3 and tested on a laptop with an Intel Core i7-1280P CPU.
%==============================
\begin{figure}[t]
  \centering
  \includegraphics[scale=0.3]{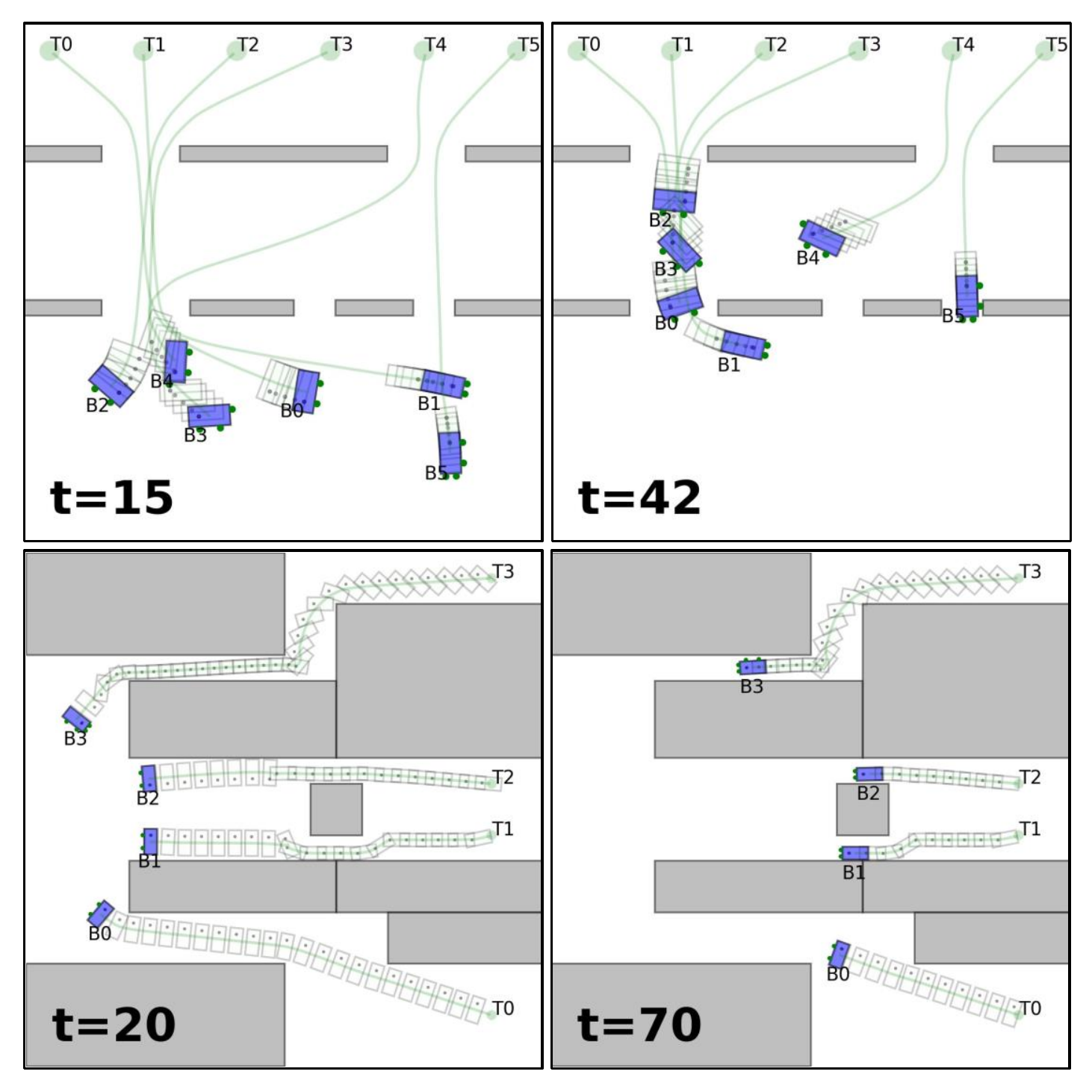}
  \vspace{-3mm}
  \caption{Snapshots of collaborative transportation within two scenarios:
  $16$ agents (blue circles) for $6$ boxes (\textbf{Top}); $10$ agents for $4$ boxes (\textbf{Bottom}).}
  \label{fig:transportation}
  \vspace{-4mm}
\end{figure}
%==============================
%==============================
\begin{figure}[t]
    \centering
    \includegraphics[scale=0.25]{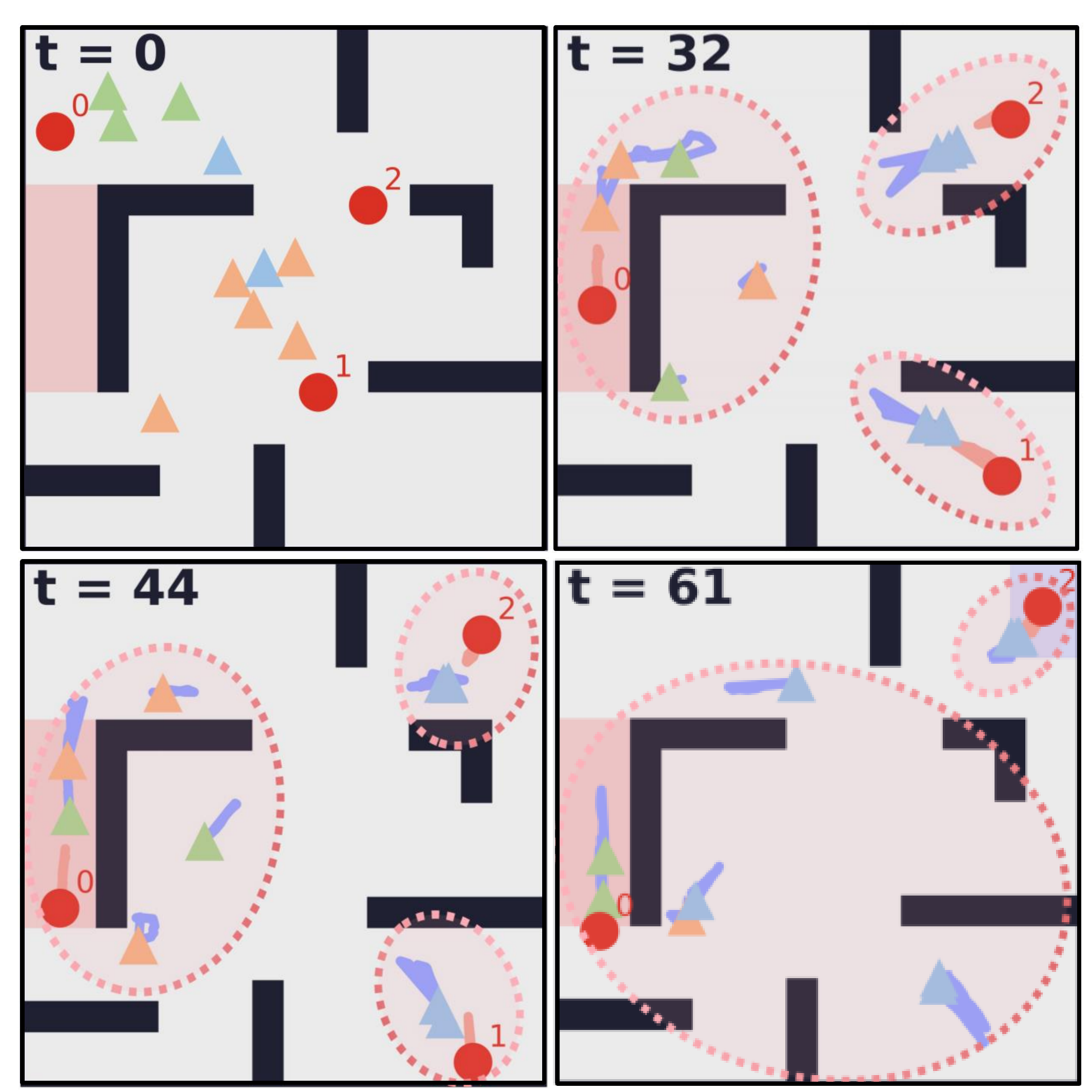}
      \vspace{-2mm}
    \caption{Snapshots of simulated dynamic capture.
      Pursuers (triangles) are encircled by their coalitions and the assigned evader (red circles).}
    \label{fig:capture}
    \vspace{-6mm}
  \end{figure}
%==============================

%==============================
\subsection{Collaborative Transportation}\label{subsec:collaborative-transport}

\subsubsection{Experiment Setup}
As introduced in Sec.~\ref{subsec:col-transport} and shown in Fig.~\ref{fig:transportation},
the workspace of size $10 m \times 10 m$ is cluttered with rectangular obstacles.
There are $16$ agents and $6$ boxes randomly distributed within the workspace.
Each box has a size of $1 m \times 0.5m$,
and the circular agents have a radius~$r=0.1 m$.
For simplicity, each robot has a first-order dynamics
and can provide bounded pushing force at specific points of the box,
thus changing the state of boxes with second-order translational invariant dynamics.
Note that the cluttered workspace introduces severe geometric constraints,
yielding a significantly more difficult transportation task than the
obstacle-free environments~\cite{tuci2018cooperative, alevizos2022bounded}.
Initially, the task cost~$f(\cdot)$ in~\eqref{eq:general-form} is
estimated by the sum of Euclidean distances from agents to the targeted box,
and the distance to its goal position.
The heuristic function $h^{\mathcal{G}}(x)$ is designed as the minimum distance from state $x$ to $\mathcal{X}_{G_m}$,
estimated by the $A^\star$ search,
and $h^{\mathcal{L}}(x)$ is designed as the kinematic cost under the local geometry constraints
from current state $x$ to next intermediate region $\hat{\mathcal{X}}$ on the shortest path.
% A low-level MPC controller is designed for each agent to provide an exact force $F \in [0,\, F_{\max}]$
% to ensure that the box moves along the planned trajectory in each mode.
%% To generate the expected trajectory from time $t$ to $t+\Delta t$,
%% nth order polynomials $P_n(t)$ are typically used to fit the expected trajectory,
%% with the parameter $\rho$ defined as a coefficient vector of the polynomials,
%% e.g. $\widehat{x}(t)=P_n^x(t)=\sum_{i=0}^n (\rho_i^x t^i)$.

\subsubsection{Results}
% \todo{Add description of the results of initial task assignment,
%   how the HO works out, how do you choose the parameters.
% }
Evolution of the system state is shown in Fig.~\ref{fig:transportation}.
Initially, the coalition formation for $6$ tasks and $16$ agents takes $9 s$,
during which~$16$ hybrid search problems are solved in the hybrid optimization layer.
It can be seen that different modes have a significant influence on overall cost.
e.g., box $5$ reaches its goal location faster
by choosing a short-side pushing mode with $2$ agents to pass through a narrow passage;
$3$ agents are assigned to box $4$ such that it can be pushed
through~$2$ passages that are far away;
box $5$ is pushed to its target by $4$ agents via almost a straight line.
To further validate the applicability, another complex scenario is considered
and shown in Fig.~\ref{fig:transportation},
where numerous sharp turns are required for the boxes to arrive at the destinations.
Consequently, more frequent mode switchings can be found in the final solution,
e.g., the coalition of $4$ agents associated with box $3$ switches the mode
from the ``long side'' to the ``short side'',
such that the box could pass through the first passage at $t =45s$.
However, since pushing from a narrow side introduces higher kinematic uncertainty
and thus larger cost,
the same coalition switches to the ``diagonally pushing'' mode
after passing through the second passage at $t=92s$.
Furthermore, for both boxes~$1$ and $2$, two agents switch
from ``long-side'' to ``short-side'' for the middle passage at $t=70$.

%==============================
\subsection{Dynamic Capture}\label{subsec:exp-capture}
%*******************
\subsubsection{Experiment Setup}
As introduced in Sec.~\ref{subsec:dynamic-capture} and shown in Fig.~\ref{fig:capture},
the workspace contains randomly placed obstacles
of various shapes like corridors and corners,
yielding a much more difficult capture task than the obstacle-free environments~\cite{pierson2016intercepting, chen2016multiplayer}.
There are $10$ pursuers and $3$ evaders randomly distributed in a workspace of size $2.5 m \times 2.5 m$.
For simplicity, all agents satisfy the single-integrator dynamics,
with the same maximum velocity of~$4m/s$.
The system state is available at all time for the pursuers.
The capture radius of pursuers is set to~$0.15m$.
As described in Sec.~\ref{subsec:dynamic-capture}, the pursuers follow three parameterized modes to capture the evaders
in a collaborative way.
% Regarding the first layer for coalition formation,
% an initial assignment $\nu_0$ is obtained by assigning each pursuer the nearest evader.
The task cost~$f(\cdot)$ in~\eqref{eq:general-form} is estimated
by the weighted average distance from the pursuers in a coalition to the assigned evader.
% Moreover, regarding the second layer for hybrid optimization,
% the vertex~$\nu$ includes the positions of all agents.
Heuristics $h^{\mathcal{G}}(\cdot)$ is designed as the minimum distance to the evader,
and $h^{\mathcal{L}}(\cdot)$ is designed according to specific local environments,
e.g.,the area of the corner envelop for enclosure mode.
Lastly, the parameter associated with mode of ``pure pursuit'' is determined solely
by the position of the targeted evader;
the parameter for the mode of ``hide and attack'' is sampled from a set of potential locations
given the workspace layout and the targeted evader;
the parameter for the mode of ``enclosure'' is determined by maximizing the region of advantage \cite{8453470}.

% ===============================
\begin{figure}[t!]
    \centering
    % ===============
    \includegraphics[width=0.6\linewidth]{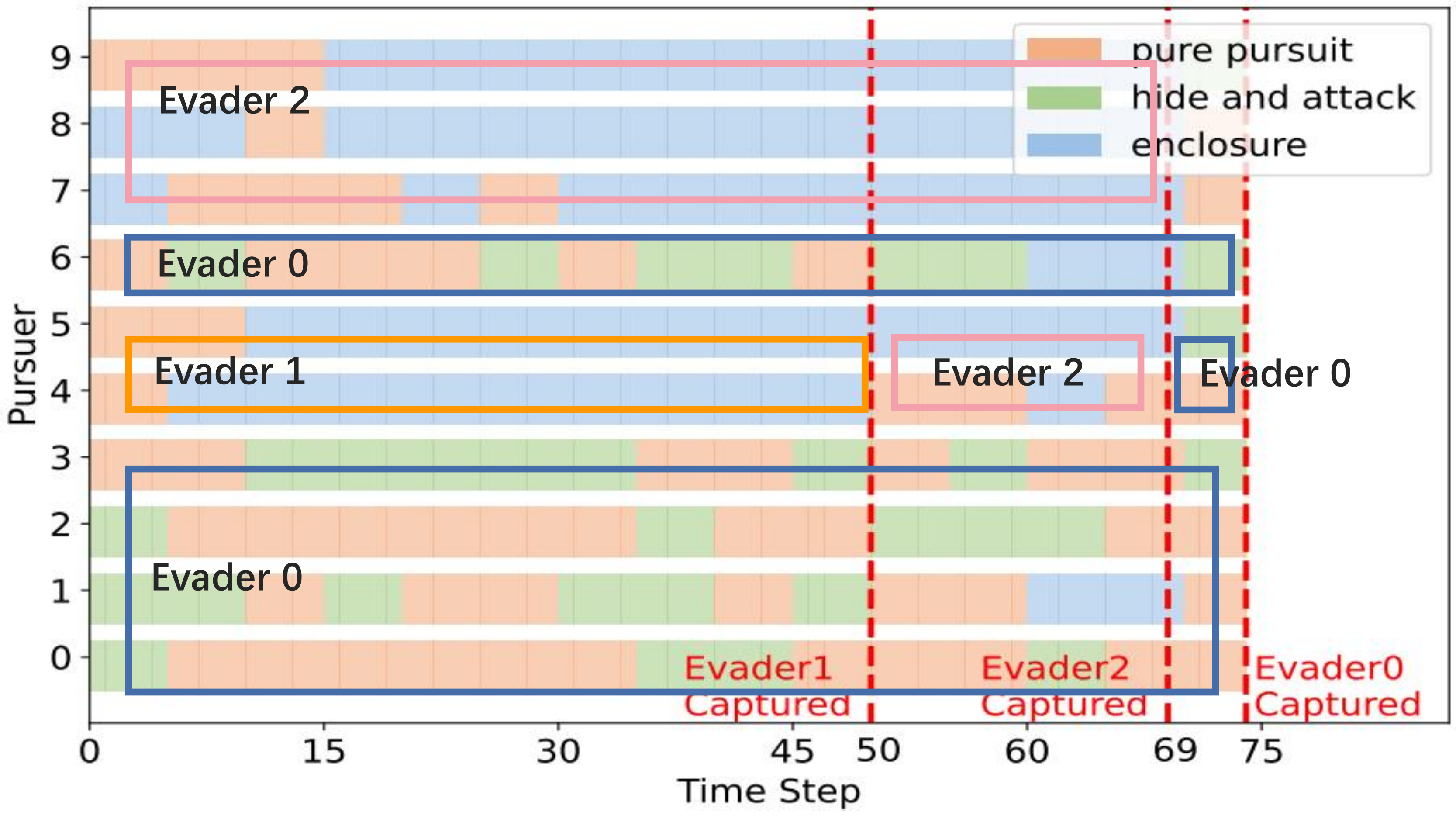}
    \vspace{-3mm}
    \caption{Evolution of task and modes for dynamic capture during one run.}
    \label{fig:mode_evolution}
    \vspace{-3mm}
\end{figure}

%*******************
\subsubsection{Results}
    The first layer of task allocation process is executed every $15$ time steps,
    while the layer of hybrid optimization is updated every $5$ time steps
    to adapt to the fast movement of evaders.
    Evolution of the system under one solution is visualized in Fig.~\ref{fig:capture}.
    Initially, the position of pursuers and evaders are initialized randomly.
    As shown in Fig.~\ref{fig:mode_evolution},
    pursers~$0,1,2,3$ and $6$ are assigned to capture evader~$0$ initially,
    while pursers $4,5$ for evader~$1$, and pursers $7,8,9$ for evader~$2$.
    At $t = 50$, evader $1$ is captured under the mode ``enclosure''
    by setting the parameters as even dividers on the boundary of advantage region.
    Afterwards, pursuers $4,5$ join the coalition for evader~$2$
    with the mode ``pure pursuit''.
    Then, evader~$2$ is captured at $t = 69$ under the mode ``enclosure'',
    after which pursuers $4, 5$ join the coalition for evader $0$
    with the mode ``hide and attack''.
    It can be seen that the selected parameters for this mode are mainly located around
    the four corners of obstacles in the middle.
    All evaders are captured at~$t=75$, during which there are in total $3$ switches of tasks
    and $24$ switches of modes.
%*******************
\subsection{Comparisons}

Effectiveness of the proposed method for the case of object transportation is
compared against two baselines:  (i) the Greedy Assignment (GA),
where each task is assigned to one closest free agent squentially
until each agent has an assigned task;
(ii) Fixed Mode (FM) where all coalitions follow only one mode,
and the layer of coalition formation remains the same.
The main metrics to compare are the sum of the cost of the hybrid plans for all coalitions
and the time when all tasks are completed.

As summarized in Table~\ref{tab:result},
for the task of collaborative transportation,
our algorithm surpasses both baselines in every metric,
e.g., the GA and FM methods have an average completion time of $127.8s$ and $132.7s$,
which is much higher than~$108.5s$ of our method.
The FM method takes considerably more time than ours in this case,
as the switching of different modes is essential for the task completion.
Secondly, for the task of dynamic capture,
it takes around $74$ time steps for our method to capture all evaders,
with the minimum and maximum capture time being $70$ and $80$ steps.
In comparison, the FM method has a much longer capture time of $108$ steps,
while the GA method takes even longer with $186$ steps in average.
It is interesting to notice that under the GA method subgroups of pursuers often
target the same evader that are close
and ignore other evaders, causing an unbalanced assignment.

% ===============================
\begin{table}[t!]
    \centering
    \scalebox{0.9}{
    \begin{tabular}{|c|c|c|c|}
    \toprule
    \midrule
    \textbf{Case} & \textbf{Method} & \makecell{\textbf{Average Time} \\ \textbf{(sec)}} & \makecell{\textbf{Mean} \\ \textbf{Cost}} \\
    \midrule
    \multirow{3}*{\makecell{\textbf{Collaborative} \\ \textbf{Transportation}}} & GA & 127.8 & 20.6\\
    & FM & 132.7 & 26.1\\
    & \textbf{Ours} & $\mathbf{108.5}$ &$\mathbf{19.0}$ \\
    \midrule
    \multirow{3}*{\makecell{\textbf{Dynamic} \\\textbf{Capture}}} & GA & 186 & $10.7$ \\
    & FM & 108 & $6.0$\\
    & \textbf{Ours} & \textbf{74} & $\mathbf{5.7}$\\
    \midrule
    \bottomrule
    \end{tabular}
    }
    \caption{Comparison with two baselines.}
    \label{tab:result}
    \vspace{-10mm}
\end{table}
% ===============================

% As summarized in Table~\ref{tab:result},
% our method can achieve 74 time steps at average of full capture for all trials,
% of which the minimum and maximum capture time is given by 70, 80.
% In comparison, the PO method achieves only an average of 108 time steps,
% with the range of capture time being 94,120.
% Lastly, the GA method has the average time of 186 and the minimum capture time is 98.
% It is interesting to see that under the GA method the pursuers often target the same evader that are close-by
% and ignore other evaders, resulting in an unbalanced assignment.
% In addition, the benefits of switching and mixing different modes during task execution is apparent
% in comparison with the PO method.

%% \iffalse
%% \begin{table}[!htb]
%% \centering
%% \caption{table}
%% \label{tab:capture-result}
%% \scalebox{0.5}{
%% \begin{tabular}{|c|c|c|c|}
%% \toprule
%% \textbf{Method} & \textbf{RandomSeed} & \makecell{\textbf{Captured Time} \\ \textbf{(sec)}} & \makecell{\textbf{Average Time} \\ \textbf{(sec)}} \\
%% \midrule
%% \multirow{5}*{Greedy Algorithm} & 1 & 103 & \multirow{5}*{186} \\
%% & 2 & 124 & \\
%% & 3 & 360 & \\
%% & 4 & 98  & \\
%% & 5 & 245 & \\
%% \midrule
%% \multirow{5}*{Pure Pursuit} & 1 & 94 & \multirow{5}*{108} \\
%% & 2 & 115 & \\
%% & 3 & 106 & \\
%% & 4 & 107  & \\
%% & 5 & 120 & \\
%% \midrule
%% \multirow{5}*{\textbf{Ours}} & 1 & 74 & \multirow{5}*{$\mathbf{74}$} \\
%% & 2 & 72 & \\
%% & 3 & 76 & \\
%% & 4 & 80  & \\
%% & 5 & 60 & \\

%% \bottomrule
%% \end{tabular}
%% }
%% \end{table}
%% \fi

%%========================================
\section{Conclusion} \label{sec:conclusion}
This work proposes a combinatorial-hybrid optimization (CHO)
framework for multi-agent systems under collaborative tasks.
The proposed approach solves simultaneously the coalition formation
and the hybrid optimization of collaborative behaviors, with a provable quality guarantee.
Future work involves the consideration of uncertain environments.

%%========================================

%========================================
\bibliographystyle{IEEEtran}
\bibliography{contents/references}

\end{document}